\documentclass[12pt,journal,compsoc]{IEEEtran} 
 
\usepackage{graphicx}
\usepackage{subfigure}
\usepackage{amsmath}
\usepackage{xcolor}
\usepackage{color, soul}

\ifCLASSOPTIONcompsoc
\else
\fi
\ifCLASSINFOpdf
\else
\fi

\begin{document}
%
\title{A dataset for Computer-Aided Detection of Pulmonary Embolism in CTA images}

\author{Mojtaba~Masoudi, Hamidreza~Pourreza, Mahdi~Saadatmand~Tarzjan, Fateme~Shafiee~Zargar, Masoud~Pezeshki~Rad, Noushin~Eftekhari
\IEEEcompsocitemizethanks{\IEEEcompsocthanksitem M. Masoudi is M.Sc. in Artificial Intelligenc, Department of Computer Engineering, Ferdowsi University of Mashhad, Iran. (e-mail: m.masoudi@mail.um.ac.ir).
\IEEEcompsocthanksitem H. Pourreza is Professor of Computer Engineering Department, Ferdowsi University of Mashahd, Mashhad, Iran.
\IEEEcompsocthanksitem M. Saadatmand Tarzjan Assistant Professor of Electrical Engineering Department, Ferdowsi University of Mashahd, Mashhad, Iran.
\IEEEcompsocthanksitem F. Shafiee Zargar is Resident of Radiology Department, Mashhad University of Medical Sciences, Mashhad,      Iran.
\IEEEcompsocthanksitem M. Pezeshki Rad is Associate Professor of Radiology Department, Mashhad University of Medical Sciences, Mashhad, Iran.
\IEEEcompsocthanksitem N. Eftekhari is M.Sc. in Artificial Intelligenc, Department of Computer Engineering, Ferdowsi University of Mashhad, Iran. }}

\IEEEtitleabstractindextext{%
\begin{abstract}
Todays, researchers in the field of Pulmonary Embolism (PE) analysis need to use a publicly available dataset to assess and compare their methods. Different systems have been designed for the detection of pulmonary embolism (PE), but none of them have used any public datasets. All papers have used their own private dataset. In order to fill this gap, we have collected 5160 slices of computed tomography angiography (CTA) images acquired from 20 patients, and after labeling the image by experts in this field, we provided a reliable dataset which is now publicly available.  In some situation, PE detection can be difficult, for example when it occurs in the peripheral branches or when patients have pulmonary diseases (such as parenchymal disease). Therefore, the efficiency of CAD systems highly depends on the dataset. In the given dataset, 66\% of PE are located in peripheral branches, and different pulmonary diseases are also included.
\end{abstract}

}

\maketitle

\IEEEdisplaynontitleabstractindextext

\IEEEpeerreviewmaketitle

\section{Introduction}
\IEEEPARstart{P}{ulmonary} embolism (PE) is a sudden blockage of a lung artery by a clot appears in an artery. This is usually caused by a blood clot in the veins of the pelvis and carry through blood flow from the heart to the lungs. This phenomenon led to the closure of the pulmonary artery, thus reducing the ability of respiratory \cite{Epidemiology} . Because of a more than 50\% artery-blockage which caused by clot, patient may die immediately, and a smaller clots can result in excessive bleeding inside the lungs. Therefore PE is a common disorder with high morbidity and mortality, hence an early and exact detection is required. Contrast-enhanced computed tomography X-ray images have been widely used in the diagnosis of PE \cite{Remy1999}. They called computed tomography angiography (CTA) images. These images have low-risk and proper display of lesions in blood vessels. Dye is dissolved in the blood and increases the contrast of vessels as a bright area in the images. However, it does not dissolve to embolism so PE appears as a dark area in CTA images \cite{Hartmann2002}. The identification of dark spots corresponding embolism by radiologists is difficult, unreliable and time-consuming, for example different radiologist may recognize different masses.
 In recent years to support radiologists and improve their performance on the challenge of pulmonary angiography in CTA images, computer-aided detection (CAD) systems are developed \cite{Masutani2002}\cite{henri2009}\cite{park2011}\cite{OZKAN2014}. The capturing time is important because it should be done when the dye is in arteries. Arteries have high contrast in return embolism and veins have low contrast in an acceptable CTA. There are other areas that are similar to emboli such as lymphatic tissue and parenchymal tissue disease. Also, partial volume effect produces similar areas on the boundaries. These values increased FP error in most of the methods that have been proposed. Therefore, dataset has a significant role in these methods.

\section{Dataset}
 A lot of research to identify PE has been done, but their efficiency is dependent on the dataset. For example, the presence or absence of various diseases of the lung or PE in the major or peripheral branches can affect the performance of designed system \cite{henri2009}. So far, there is no publicly available dataset for PE, so each paper has used their own private dataset.  In order to Verify and compare these methods, a public dataset must be used. We have provided and released a public dataset that can be downloaded in DICOM format \cite{addressMVlab}.
The gold-standard annotation is done by two radiologists named F.Shafiee (a board certiﬁed radiologist with over 5 years of experience reading CTA) and M.Pezeshki (head of the radiology unit of Emam-Reza Hospital in Mashhad, Iran with more than 18 years of clinical experience). F.Shafiee first analyzed images and then marked the region of interest (ROI), after that M.Pezeshki re-examined ROI. From these markings, a semi-automated method for PE segmentation was applied. This PE segmentation is consists of a thresholding step based on Hounsﬁeld units, followed by a morphological operation and connected component analysis. Each segmentation is then manually inspected to remove spurious pixels. The gold-standard is given as segmentation masks with 0 value for background and 1 for the PEs. As mentioned before PE position affects the performance of the CAD systems, so in this dataset, we have selected cases which have PE in different branches. In Table \ref{tab:table2}, we show how many PE is in the major and peripheral artery. As can be seen, a high percentage of PE (66\%) is in the peripheral artery which is a better case to have more PE in the peripheral of the artery branches. There are also some types of pulmonary diseases in some cases.

	\begingroup
	\begin{table}
		\caption {\label{tab:table2} summary of the pulmonary embolism distribution.} 
			\begin{tabular}{ccccc}
				Case & \#PE &\multicolumn{2}{c}{PE} & All	PE \\ 
				\cline{3-4}  
				&&  Major~Artery  & Peripheral~Artery                         &             
                                 \\    \hline 
				$Patient01$ & 201 & 46201  & 34  & 80 \\
				$Patient02$ & 185 & 0   & 21  & 21 \\
				$Patient03$ & 210 & 39  & 55  & 94 \\
				$Patient04$ & 197 & 17  & 29  & 46 \\
				$Patient05$ & 217 & 71  & 92  & 163 \\
				$Patient06$ & 232 & 0   & 15  & 15 \\
				$Patient07$ & 197 & 95  & 147 & 242 \\
				$Patient08$ & 273 & 23  & 36  & 59 \\
				$Patient09$ & 237 & 0   & 53  & 53 \\
				$Patient10$ & 204 & 0   & 8   & 8 \\
				$Patient11$ & 217 & 0   & 18  & 18 \\
				$Patient12$ & 178 & 0   & 77  & 77 \\
				$Patient13$ & 189 & 30  & 23  & 53 \\
				$Patient14$ & 217 & 0   & 98  & 98 \\
				$Patient15$ & 250 & 15  & 31  & 46 \\
				$Patient16$ & 235 & 18  & 59  & 77 \\
				$Patient17$ & 475 & 194 & 60  & 254 \\
				$Patient18$ & 452 & 54  & 102 & 156 \\
				$Patient19$ & 370 & 51  & 300 & 351 \\
				$Patient20$ & 424 & 19  & 48  & 67 \\
				\hline 
				$All$ & 5160 & 672(34\%) & 1306(66\%) & 1978  
	\end{tabular}
	\end{table}

%

\ifCLASSOPTIONcompsoc

\ifCLASSOPTIONcaptionsoff
  \newpage
\fi

\bibliographystyle{IEEEtran}
\bibliography{bibliography}

%
%

\end{document}